\documentclass[preprint,12pt]{elsarticle}

\usepackage[utf8]{inputenc}  
\usepackage[T1]{fontenc}     
\usepackage[english]{babel}  

\usepackage{amsmath,amssymb} 
\usepackage{physics}         

\usepackage{graphicx}        
\usepackage[dvipsnames]{xcolor}  
\usepackage{lineno}          
\usepackage{float}           
\usepackage{lscape}          
\usepackage{rotating}        
\usepackage{longtable}       
\usepackage[numbers,sort&compress]{natbib}  
\usepackage{booktabs}
\usepackage{array}
\usepackage{booktabs}

\biboptions{sort&compress}

\begin{document}

\begin{frontmatter}

\title{Integrating Complexity and Biological Realism: High-Performance Spiking Neural Networks for Breast Cancer Detection}

\author[label1]{Zofia Rudnicka}
\author[label1]{Janusz Szczepanski}
\author[label1]{Agnieszka Pregowska\corref{cor1}}
\cortext[cor1]{Corresponding author: aprego@ippt.pan.pl}

\address[label1]{Institute of Fundamental Technological Research, Polish Academy of Sciences,\\
Pawinskiego 5B, 02--106 Warsaw, Poland}

\begin{abstract} 
Spiking Neural Networks (SNNs) are neuromorphic models inspired by biological neurons, where information is transmitted through discrete spike events. Their event-driven nature enables efficient encoding of spatial and temporal features, making them suitable for dynamic time-dependent data processing. Despite their biological relevance, SNNs have seen limited application in medical image recognition due to difficulties in matching the performance of conventional deep learning models. To address this, we propose a novel breast cancer classification approach that combines SNNs with Lempel-Ziv Complexity (LZC) a computationally efficient measure of sequence complexity. LZC enhances the interpretability and accuracy of spike-based models by capturing structural patterns in neural activity. Our study explores both biophysical Leaky Integrate-and-Fire (LIF) and probabilistic Levy-Baxter (LB) neuron models under supervised, unsupervised, and hybrid learning regimes. Experiments were conducted on the Breast Cancer Wisconsin (Diagnostic) dataset using numerical features derived from medical imaging. LB-based models consistently exceeded 90.00\% accuracy, while LIF-based models reached over 85.00\%. The highest accuracy of 98.25\% was achieved using an ANN-to-SNN conversion method applied to both neuron models comparable to traditional deep learning with back-propagation, but at up to 100 times lower computational cost. This hybrid approach merges deep learning performance with the efficiency and plausibility of SNNs, yielding top results at lower computational cost.
We hypothesize that the synergy between temporal-coding, spike-sparsity, and LZC-driven complexity analysis enables more-efficient feature extraction. Our findings demonstrate that SNNs combined with LZC offer promising, biologically plausible alternative to conventional neural networks in medical diagnostics, particularly for resource-constrained or real-time systems.

\end{abstract}


\end{frontmatter}


\section{Introduction}
Breast cancer represents the most frequently diagnosed cancer among women in the United States, excluding skin cancers. It constitutes approximately 30.00\% of all newly diagnosed cancers in women each year \cite{ACS2025}. In Europe, breast cancer is similarly prevalent. It is the most commonly diagnosed cancer among women and a leading cause of cancer-related mortality. This makes it an important diagnostic area, especially in the context of incorrect identification \cite{ThomassinNaggara2024,Bahreyni2024}.
\par
Recently, Artificial Intelligence (AI) is becoming an increasingly important support for doctors in the diagnostic process, in particular various type of cancer \cite{Gerratana2025}. Their design is often a balance between theoretical rigor and practical applicability. While mathematics and statistics provide the foundation, translating these into algorithms that can operate on vast and diverse datasets requires creative programming skills. For example, in the study \cite{Lu2022,Xiong2025} proposed the diagnosis of breast cancer using deep neural networks that consist of pre-trained ResNet-18 and three recurrent neural networks based on perceprons. In fact, the limitations of classical AI, based on perceptrons contribute to the discovery of new possibilities of neural network models to increase the speed of computation. Leading candidates to overcome the limitations of ANN, an energy-efficient alternative are spiking neural networks (SNNs)\cite{GhoshDastidar2009,Lin2016, Huang2022,Dampfhoffer2023}. However, the training of the SNNs due to quite complicated dynamics and the non-differentiable nature of the spike activity remains a challenge.

\par
Despite growing application field of SNN in medicine, there is limited research in the current literature on the application of SNNs for breast cancer diagnosis \cite{Sangeetha2015,Javanshir2023,Das2025}. In \cite{Fu2022}, ReSuMe learning algorithm was applied to SNN architecture based on a Liquid State Machine (LSM), optimized using the Fruit Fly Optimization Algorithm (FOA) to effective breast cancer image recognition. The proposed method was evaluated on three medical image datasets: BreastMNIST, mini-MIAS, and BreaKHis. SNNs using linear time encoding, entropy-based time encoding, silence-based encoding, as well as an improved SNN configuration, achieved classification accuracy above 90.00\%. In comparison, the original SNN configuration achieved an accuracy of approximately 80.00\%. On the other hand, \cite{Fu2022} the same architecture was combined with You Only Look Once (YOLO), and applied to classification of breast lesions with ultrasound and X-ray datasets, achieving the classification accuracy over 90.00\%. The study \cite{Heidarian2024} also focuses on improving learning algorithms for SNNs in the context of breast cancer diagnostics. It introduces a novel temporal feedback backpropagation method. In contrast, \cite{Sboev2018} explores the use of Spike Timing Dependent Plasticity (STDP), achieving an accuracy of 96.00\%. Another noteworthy approach to applying SNNs in breast cancer diagnostics is proposed in \cite{Tang2025}. Furthermore, \cite{Fu2022} combines the STDP learning rule with gradient descent mechanisms. In addition, \cite{Wade2010} introduces Synaptic Weight Association Training (SWAT) for SNNs, achieving an accuracy exceeding 95.00\%. All of the aforementioned approaches employ the Leaky Integrate-and-Fire (LIF) neuron model. In contrast, the study \cite{Liu2024} considers an alternative neuron model for breast cancer diagnosis, namely the Nonlinear Synaptic Nonlinear Processing – Adaptive Update (NSNP-AU) neuron. This model incorporates nonlinear dynamics not only at the synaptic level but also within the neuron's internal processing mechanisms. On the other hand, an interesting approach for breast cancer recognition was proposed by \cite{Saranirad2021}, namely Degree of Belonging SNN (DoB-SNN). It operates as a two-layer network that utilizes the degree of belonging to determine class membership, offering a nuanced classification mechanism that moves beyond binary decisions. 
\par
In this study, we propose a novel approach for breast cancer classification that combines spiking neural networks (SNNs) with Lempel-Ziv Complexity (LZC) \cite{Ziv1976}. The network architecture was based on commonly used Leaky Integrate-and-Fire (LIF) as a reference and probabilistic Levy-Baxter (LB) neuron model. To the best of our knowledge, the LB model has not previously been employed in the construction of SNNs for cancer diagnostics. By leveraging the temporal precision and biological plausibility of SNNs alongside the ability of LZC to quantify the structural complexity of spike patterns, this hybrid method enables efficient, interpretable, and noise-robust classification of spatiotemporal neural data. The approach is particularly effective for signals with variable temporal dynamics, such as Poisson-distributed spike trains.
To validate our method we have considered also different learning algorithms. Our pilot study suggests that proposed approach can be used as a clinical tool to successfully classify breast cancer. This is a promising direction in the classification of such a complex diseases like breast cancer \cite{Fu2022,Tang2025}.

\section{Basics Notation}
In Table \ref{tab:notation} the notation used is presented.

\section{Neuron Models}
\begin{figure}[ht] 
\centering
    \includegraphics[width=1\linewidth]{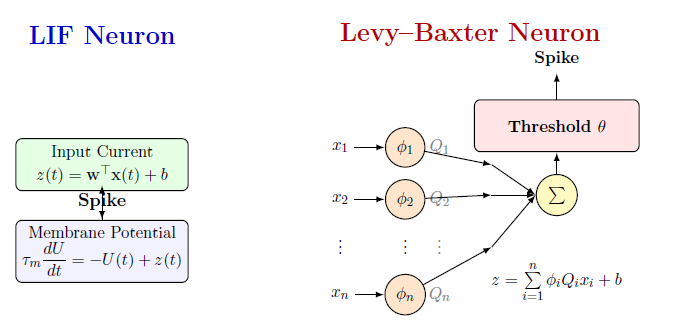}
\caption{Schematic of the LIF and Levy--Baxter neuron models.}
\label{fig:neurons}
\end{figure}

In this study, we consider as SNN building blocks Leaky Integrate-and-Fire and Levy-Baxter neurons models. The principle of operation of both neurons is shown in Figure \ref{fig:neurons}.

\begin{table}[H]
\centering
\scriptsize
\renewcommand{\arraystretch}{1.05}
\begin{tabular}{l p{5.8cm} l}
\toprule
\( \mathbf{x} = [x_1, x_2, \dots, x_n] \in \mathbb{R}^n \) & Input vector & Input \\
\( \mathbf{w} = [w_1, w_2, \dots, w_n] \in \mathbb{R}^n \) & Weight vector & Parameters \\
\( b \) & Bias & Parameters \\
\( \theta \) & Threshold & Parameters \\
\( z = \mathbf{w}^T \mathbf{x} + b \) & Weighted sum & Computation \\
\( f(z) \) & Activation function & Neuron Model \\
\( \tau_m^+ \in \mathbb{R}^+ \) & Membrane time constant & Neuron Model \\
\( R_m \in \mathbb{R}^+ \) & Membrane resistance & Neuron Model \\
\( I(t) \in \mathbb{R} \) & Input current at time \( t \) & Dynamics \\
\( V_m(t) \in \mathbb{R} \) & Membrane potential at time \( t \) & Dynamics \\
\( t_f \in \mathbb{R}^+ \) & Firing time & Output \\
\( S_i(t) \in \{0, 1\} \) & Spike train (binary) & Output \\
\( t_i^k \in \mathbb{R}^+ \) & Spike times & Output \\
\( \Theta(V_n) \in \{0,1\} \) & Heaviside function & Nonlinearity \\
\( \eta \in \mathbb{R}^+ \) & Learning rate & Learning \\
\( \eta_{+}, \eta_{-} \in \mathbb{R}^+ \) & STDP time constants & Learning \\
\( A_{+}, A_{-} \in \mathbb{R}^+ \) & STDP amplitudes & Learning \\
\( t_{\text{pre}}, t_{\text{post}} \in \mathbb{R}^+ \) & Pre- and postsynaptic spike times & Learning \\
\( t_{\text{spike}} \in \mathbb{R}^+ \) & Spike timing & Dynamics \\
\( S_{\text{pre}}(t), S_{\text{post}}(t) \in \{0,1\} \) & Pre/post binary spike trains & Learning \\
\( K(t) \) & Kernel function & Learning \\
\( t_i^{\text{target}}, t_i^{\text{actual}} \) & Target/actual spike times & Learning \\
\( E \in \mathbb{R}^+ \) & Error function & Learning \\
\( \frac{\partial E}{\partial w_i} \) & Error gradient & Learning \\
\( \Phi(t_{\text{pre}}, t_{\text{post}}) \) & Loss function & Learning \\
\( S_{\text{teach}}(t) \) & Teacher spike train & Learning \\
\( w_{\text{SNN}}, w_{\text{ANN}} \) & Weights in SNN/ANN & Parameters \\
\( \tau_{\text{syn}} \in \mathbb{R}^+ \) & Synaptic time constant & Learning \\
\( r(t) \in \mathbb{R} \) & Reward signal & Reinforcement \\
\( \Delta w_{\text{STDP}} \) & STDP-based weight change & Learning \\
\( U(w_i) \) & Uncertainty function & Meta-learning \\
\bottomrule
\end{tabular}
\caption{Summary of mathematical notation used in the study.}
\label{tab:notation}
\end{table}

The Leaky Integrate-and-Fire neuron model provides a widely used in SNN framework for describing the subthreshold dynamics of a neuron's membrane potential $U(t) \in \mathbb{R}$ \cite{Dayan2000,Dutta2017,Yao2022,Huang2024}. Given an input vector $x \in \mathbb{R}^{n}$, weight vector $w \in \mathbb{R}^{n}$, and bias $b \in \mathbb{R}^{n}$, the total input current can be express as
\begin{equation}
    I(t)=w^{T}\textbf{x}(t)+b=z(t)
\end{equation}
where $z(t) \in \mathbb{R}^{n}$ represents the weighted sum at time $t$. The membrane potential evolves according to the differential equation
\begin{equation} 
\tau_m \frac{dU(t)}{dt} = -U(t) + z(t). \label{eq:LIF_table} 
\end{equation}
When the potential $U(t)$ reaches a predefined threshold $U_{th}$, a spike is generated, and the potential is reset to a lower resting value $U_{r}$. This mechanism models the accumulation and leakage of charge across the membrane, mimicking basic neuronal firing behavior.

The Levy–Baxter neuron model introduces a probabilistic mechanism to simulate variability in synaptic transmission and response \cite{Levy1996,Paprocki2020,Klamka2020,Paprocki2024}. The input to the neuron is defined as a vector $x=[x_{1},x_{2},…,x_{n}]$, where each $x_{i}$ is a binary signal that represents the presence or absence of the input of the $i$-th synapse. The synaptic response is modulated by two independent sources of stochasticity: a Bernoulli distributed random variable $\phi_{i}$, representing the probability of neurotransmitter release with success probability $s \in [0, 1]$, and the amplitude is scaled by a random variable $Q_i$ uniformly distributed over $[0, 1]$. The total synaptic excitation is
\begin{equation}
    \sigma = \sum_{i=1}^n \phi_i Q_i x_i
\end{equation}

and the neuron's output is given by the threshold rule
\begin{equation}
    z = \begin{cases}
    1 & \text{if } \sigma \geq 0 \\
    0 & \text{if } \sigma < 0
    \end{cases}
\end{equation}

where $z=1$ indicates a spike. This probabilistic framework captures synaptic variability, with $x_{i}$ as binary inputs, $\phi_{i}$ representing the probabilistic nature of neurotransmitter release, and $Q_{i}$ modeling amplitude fluctuations.

\begin{figure}[ht]
\centering
    \includegraphics[width=1\linewidth]{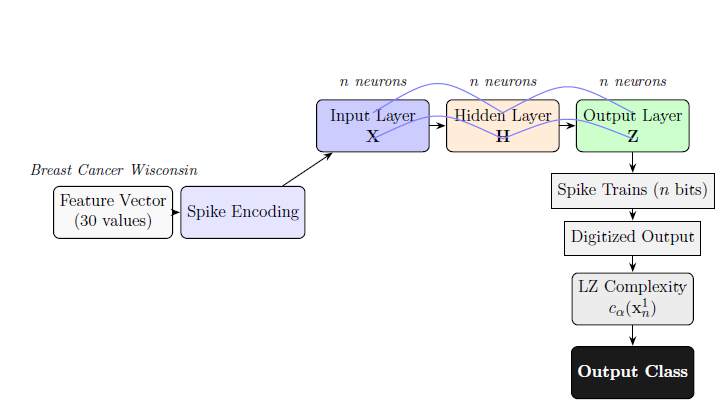}
\caption{Spiking neural network scheme inspired by biological neuron connectivity. Input feature vectors from the Breast Cancer dataset are encoded into spike trains and passed through fully connected spiking layers. The final spike output is digitized and evaluated using Lempel-Ziv Complexity for classification.}
\label{fig:task}
\end{figure}

\section{Neural Network Architecture}
The neural network under consideration consists of three layers: input, hidden, and output, each containing $n \in \{2, 8, 16, 30\}$ neurons see Figure \ref{fig:task}. Every neuron in one layer is connected to every neuron in the next layer, forming dense inter-layer connectivity. The neural dynamics are governed both by LIF model, as introduced in \cite{Dutta2017} and by Levy Baxter neuron model \cite{Levy1996}. For each learning algorithm under study, multiple hyperparameter configurations were explored, and the reported results correspond to the most efficient configuration in each case. The parameters of neurons model were systematically varied across the following ranges: membrane threshold values $\theta \in [0.1, 0.5]$,  membrane potential decay constants $\delta \in [0.01, 0.1]$, and learning rates $\eta \in [10^{-4}, 10^{-1}]$. In order to mitigate overfitting and encourage sparsity in the synaptic weight matrix $\textbf{W}=[w_{ij}]$, we incorporate a regularization term into the loss function, defined as
\begin{equation}
\mathcal{L}_{\text{reg}} = \lambda \sum_{i,j} w_{ij}^2,
\end{equation}

where $\lambda >0$ is the regularization coefficient.
\par
The network is tasked with processing binary sequences of fixed length $L=30$. Each sequence is encoded into an $n$-dimensional spike train, propagated through the network, and subsequently decoded back into a binary sequence. To evaluate the complexity of the network's output, we employ the LZ complexity measure \cite{Ziv1976,Paprocki2020,Klamka2020,Paprocki2024}, a widely accepted metric for quantifying algorithmic randomness. Let $\mathbf{x}_n^1 = [x_1, x_2, \dots, x_n]$ be a binary sequence with $ x_i \in \{0,1\} $. The LZ complexity $ C_{\alpha}(\mathbf{x}_n^1) $ quantifies the number of distinct substrings found during sequential parsing. The normalized complexity is defined as
\begin{equation} \label{Hebb}
c_{\alpha}(\mathbf{x}_n^1) = \frac{C_{\alpha}(\mathbf{x}_n^1)}{n}\log_{\alpha}{n},
\end{equation}
where  $\alpha = 2 $ for binary sequences. Asymptotically, $ c_{2}(\mathbf{x}_n^1) \to 1 $ for random sequences and $ c_{2}(\mathbf{x}_n^1) \to 0 $ for deterministic sequences.
This approach was proposed and tested on artificial signals in \cite{Shrestha2017,Alemanno2023}. The implementation is conducted in \texttt{Python}. Core dependencies include \texttt{NumPy} for numerical operations, \texttt{Scikit-learn} for data pre-processing and evaluation, and \texttt{Matplotlib} for visualization. Hyperparameter optimization is performed using \texttt{Optuna}. The preprocessing pipeline consists of feature normalization via \texttt{MinMaxScaler}, dimensionality reduction using Principal Component Analysis (PCA), and class balancing through the Synthetic Minority Over-sampling Technique (SMOTE). For the evaluation of the proposed classification task, we applied accuracy.

\begin{figure}[ht]
    \centering
    \includegraphics[width=0.7\linewidth]{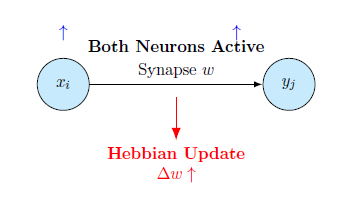}

    \caption{Schema of the operating principle of the Hebbian learning algorithm. Neurons 
    \(x_{i}\) and \(y_{y}\) represent pre- and post-synaptic neurons, respectively. When both neurons are active (indicated by simultaneous upward arrows), the synapse connecting them \(w\) undergoes potentiation.}
    \label{fig:hebbian}
\end{figure}

\section{Learning Algorithms}

\begin{figure} 
 \centering

    \includegraphics[width=0.65\linewidth]{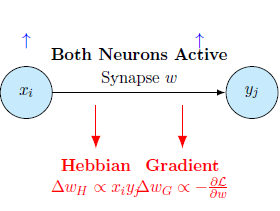}

  \caption{Schema of the operating principle of the Hebbian learning algorithm combined with gradient decent learning rule. Neurons \(x_{i}\) and \(y_{y}\) represent pre- and post-synaptic neurons, respectively. When both neurons are active (indicated by simultaneous upward arrows), Hebbian learning updates the synaptic weight proportionally to the product of their activities, reinforcing co-activation. 
  }
    \label{fig:hebbian_gradient}

\end{figure}

We have considered a wide range of different learning algorithms, including unsupervised like Hebbian, modified Hebbian (i.e. Hebbian combined with Gradient Decent) and STDP, supervised such as back propagation, tempotron learning rule, and Spike Prop as well as hybrid like reward based, Bio-inspired Active Learning (BAL) and ANN SNN Conversion. The first natural choise is bio-inspired Hebbian learning that strengthens synaptic connections when both pre- and post-synaptic neurons are active simultaneously \cite{Rathi2023}

\begin{equation} 
\Delta w_{i} = \eta , S_{\text{pre}}(t) S_{\text{post}}(t), \label{Hebb} 
\end{equation} 
 see Figure \ref{fig:hebbian}. 
Although Hebbian learning is biologically plausible and computationally simple, it tends to be less effective in deep architectures due to the absence of global error feedback \cite{Amato2019}. To address this limitation, we also explore a gradient-modified Hebbian learning rule, where local Hebbian updates are supplemented with gradient-based adjustments derived from a global loss function, see Figure \ref{fig:hebbian_gradient}.

\begin{figure} \label{stdp}
    \centering

    \includegraphics[width=1.0\linewidth]{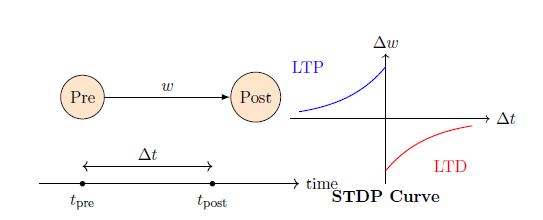}

    \caption{Schema of the operating principle of the Spike-Timing-Dependent Plasticity learning algorithm. The orange circles represent the pre-synaptic and post-synaptic neurons, respectively. The diagram depicts scenarios where the pre-synaptic neuron fires either before or after the post-synaptic neuron (timeline illustrates the firing times of both neurons, highlighting the time difference \(\Delta t\) between pre- and post-synaptic spikes). Thus, a positive value of \(\Delta t\) (pre-synaptic neuron fires before post-synaptic neuron) leads to Long-Term Potentiation (LTP), while a negative value of \(\Delta t\)  (pre-synaptic neuron fires after post-synaptic neuron) results in Long-Term Depression (LTD).}
    \label{fig:stdp}
\end{figure}

Other, unsupervised learning algorithm, which we explore is Spike-Timing Dependent Plasticity. It represents a biologically inspired learning mechanism in which synaptic modifications are determined by the precise temporal relationship between pre-synaptic and postsynaptic spikes \cite{Markram2011,Merolla2014,Chakraborty2021,Lagani2023}, see Figure \ref{fig:stdp}.  
The weight change $\Delta w_i$ is defined as:

\begin{equation}
    \Delta w_{i} =
    \begin{cases}
        A_{+}e^{-\frac{(t_{post}-t_{pre})}{\tau_{+}}} & t_{post} > t_{pre} \\
        -A_{-}e^{-\frac{(t_{post}-t_{pre})}{\tau_{-}}} & t_{pre} > t_{post} .
    \end{cases}
\end{equation}

where $A_{+}, A_{-}$ and $\tau_{+}, \tau_{-}$ are the amplitude and time-constant parameters, respectively. STDP represents a temporally asymmetric variant of Hebbian plasticity (\ref{Hebb}), enabling unsupervised learning in event-driven systems. It is particularly effective in encoding spatiotemporal patterns in sensory or dynamical input streams. However, the model is sensitive to hyperparameter choices and becomes very computationally expensive as the network size grows \cite{Cellina2023}.

\begin{figure}[ht]
    \centering

    \includegraphics[width=0.65\linewidth]{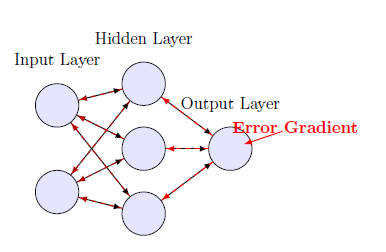}

    \caption{Schema of the operating principle of the backpropagation learning algorithm.
Blue circles represent neurons involved in the forward pass, solid arrows indicate the flow
of activations (forward pass), while dashed red arrows depict the backpropagation of error
gradients.}
    \label{fig:bp}
\end{figure}

When it comes to supervised learning algorithms, the most commonly used backpropagation consists of two main computational phases, namely forward and backward propagation, which enable efficient weight optimization via gradient descent, see Figure \ref{fig:bp} \cite{Rumelhart1986,Hameed2016,Wojcik2018,Singh2022}. The weight update in backpropagation is typically defined as

\begin{equation} 
\Delta w_{i} = -\eta \frac{\partial E}{\partial w_{i}}, 
\label{BP} \end{equation}

where $\eta$ is the learning rate, and $E$ denotes the error function. The gradient $\frac{\partial E}{\partial w_{i}}$ is computed using the chain rule as follows

\begin{equation} 
\frac{\partial E}{\partial w_{i}} = \frac{\partial E}{\partial V_{m}} \cdot \frac{\partial V_{m}}{\partial w_{i}}, 
\end{equation}
where $V_{m}$ is the input to the activation function of the $m$-th neuron \cite{Kaur2023}.

\begin{figure}[ht]
    \centering

    \includegraphics[width=0.5\linewidth]{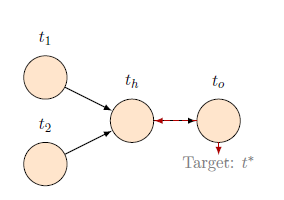}

    \caption{Scheme of the operating principle of the SpikeProp learning rule. Input neurons spike at times \(t_{1}\) and \(t_{2}\), which are integrated by a hidden neuron that spikes at \(t_{h}\). The output neuron subsequently spikes at time \(t_{0}\). The red dashed arrow denotes the timing error signal, defined as the difference between the actual spike time \(t_{0}\) and the desired target spike time \(t^{*}\).}
    \label{fig:spikeprop}
\end{figure}

In turn, SpikeProp formulates supervised learning in spiking neural networks through gradient descent on spike timing errors \cite{Shrestha2017}, see Figure \ref{fig:spikeprop}. The weight update is derived from the chain rule:

\begin{equation} \frac{\partial E}{\partial w_{i}} = \sum_{t_{post}} \frac{\partial E}{\partial t_{i}^{\text{actual}}} \cdot \frac{\partial t_{i}^{\text{actual}}}{\partial w_{i}}, \end{equation}

where $E$ denotes the error functional and $t_i^{\text{actual}}$ the observed spike times. By extending the backpropagation framework to temporal spike codes, SpikeProp facilitates gradient-based optimization in time-driven spiking models. However, this type of learning is computationally intensive and must address the inherent non-differentiability of spike events, often through surrogate gradient techniques or approximations.

\begin{figure} [h]
    \centering

    \includegraphics[width=1.0\linewidth]{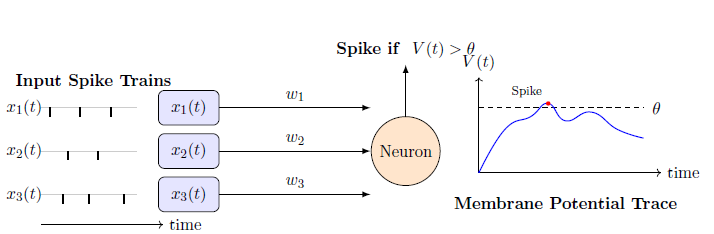}
    
  \caption{Schematic illustration of the operating principle of the Tempotron learning rule. Input spike trains \(x_i(t)\) generate time-dependent synaptic inputs to a neuron, weighted by \(w_i\). These inputs contribute to the neuron's membrane potential \(V(t)\), which integrates over time. A spike is generated if \(V(t)\) exceeds the threshold \(\theta\). Learning is achieved by modifying the weights based on the neuron’s response to the input spike pattern.}
    \label{fig:tempotron}
\end{figure}

An interesting approach is Tempotron learning rule that modifies synaptic weights to enable binary classification of spatiotemporal spike patterns \cite{Gutig2006,Urbanczik2009,Yu2014}, see Figure \ref{fig:tempotron}. Synaptic updates are given by:

\begin{equation} \Delta w_{i} = \eta \sum_{t \in T_{\text{pre}}} K(t_{f} - t), \end{equation}

where $\eta$ is the learning rate, $T_{\text{pre}}$ denotes the set of presynaptic spike times, $t_f$ is the postsynaptic firing time, and $K(\cdot)$ is the postsynaptic potential kernel. Unlike rate-based models, the Tempotron emphasizes the causal role of precise spike timing in shaping synaptic efficacy. This biologically motivated mechanism demonstrates that temporal coding can serve as a sufficient basis for learning and decision-making in spiking neural systems.

\begin{figure}[ht]
    \centering
    \includegraphics[width=1.0\linewidth]{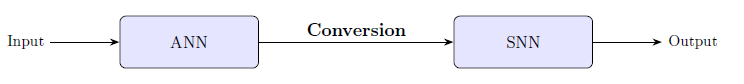}
    \caption{Schematic illustration of the operating principle of theANN-to-SNN conversion. A trained ANN is transformed into a SNN for efficient spike-based inference}
    \label{fig:ann-snn}
\end{figure}

\begin{figure} [h]
    \centering

        \includegraphics[width=0.7\linewidth]{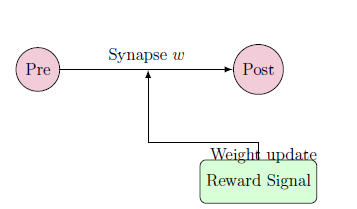}

    \caption{Schematic illustration of the operating principle of the reward-based SNN learning algorithm. Pre-synaptic neuron forms a synaptic connection \(w\) to post-synaptic neuron. A modulatory reward signal influences the synaptic plasticity rule, reinforcing or weakening the synapse based on the outcome.}
    \label{fig:rb}
\end{figure}

We also consider hybrid learning algorithms like ANN-SNN conversion enables the transfer of trained ANN weights to spiking architectures \cite{Midya2019}, see Figure \ref{fig:ann-snn}. The mapping is typically defined as:

\begin{equation} w_{\text{SNN}} = \frac{w_{\text{ANN}}}{\tau_{\text{syn}}}, \end{equation}

where $\tau_{\text{syn}}$ denotes the synaptic time constant. Activation patterns from ANNs are preserved in SNNs by frequency coding, where the firing rates approximate continuous outputs. However, network accuracy is sensitive to threshold values, with suboptimal settings producing either non-firing or bursting activity responses \cite{Yang2025}.

Other hybrid algorithm that we explore is reward-based spike-timing-dependent plasticity. It introduces a modulatory reinforcement signal to conventional STDP (\ref{stdp}) dynamics \cite{Khoei2024}, see Figure \ref{fig:rb}:

\begin{equation} \Delta w_{ij} = \eta , r(t) \left[ S_{\text{pre}}(t) * S_{\text{post}}(t) \right], \end{equation}

where $r(t)$ is a scalar reward signal, $S_{\text{pre}}(t)$ and $S_{\text{post}}(t)$ are spike trains and $*$ denotes temporal convolution. This mechanism combines unsupervised temporal plasticity with reinforcement-based modulation.

\begin{figure}[h]

    \centering
    \includegraphics[width=0.7\linewidth]{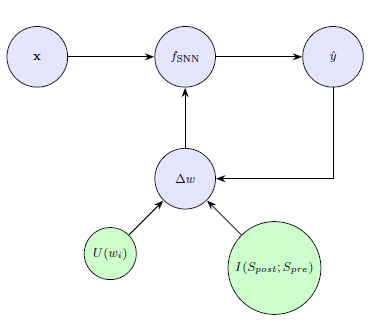}

\caption{Schematic of the BAL. Input \(x\) is processed by the network \(f_{SNN}\), producing a prediction \(\hat{y}\). The synaptic update \(\Delta w\)is driven by a plasticity rule modulated by the synaptic uncertainty \(U(w_{i})\) and the expected mutual information \(I(S_{post}; S_{pre})\) between pre- and post-synaptic spikes.}

    \label{fig:bal}
\end{figure}

We also consider Bio-Inspired Active learning that integrates synaptic plasticity with uncertainty-driven sample selection \cite{Conforth2008,Bianchi2020,Xie2024}, see Figure \ref{fig:bal}. A typical weight update rule is:

\begin{equation} \Delta w_{ij} = \eta \cdot U(w_{i}) \cdot \mathbb{E} \left[ I(S_{\text{post}}; S_{\text{pre}}) \right], \end{equation}

where $U(w_i)$ represents model uncertainty and $I(\cdot,;\cdot)$ denotes mutual information. BAL prioritizes informative samples for label acquisition, achieving efficient learning under data constraints, consistent with neuro-biological mechanisms \cite{Xie2024}.

\section{Input Dataset} 
\begin{figure}[ht]
\centering

    \includegraphics[width=1.0\linewidth]{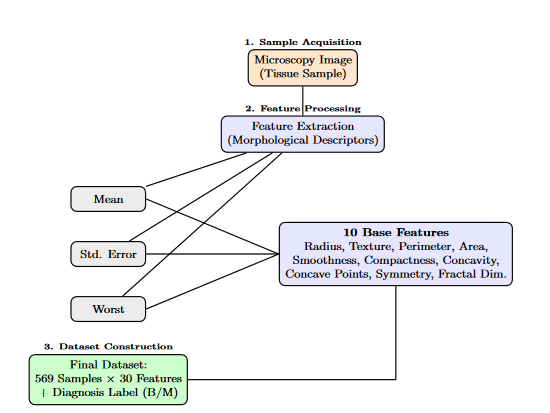}
\caption{Construction of the Breast Cancer Wisconsin (Diagnostic) dataset.}
\label{database}
\end{figure}

In this study, we used the Breast Cancer Wisconsin (Diagnostic) dataset, available through Scikit-learn, which is a widely used benchmark for binary classification tasks. The construction of this database was presented in Figure \ref{database}. It comprises 569 samples with 30 numerical features derived from digitized images of fine needle aspirates (FNA) of breast masses. Each sample is labeled either malignant or benign, based on clinical diagnosis. These features encapsulate various characteristics of the cell nuclei present in the images, including radius, texture, perimeter, area, smoothness, compactness, concavity, concave points, symmetry, and fractal dimension. Each sample in the dataset is labeled based on clinical diagnosis: 212 cases are identified as malignant, and 357 as benign. This data set is frequently used in machine learning research to evaluate diagnostic models in medical imaging and cancer classification. A standard train–test split of 80\% for training and 20\% for testing was employed.

\section{Results and Discussion} 

\begin{table}[H] 
\centering
\caption{Comparison of classification accuracy across different learning algorithms, considering the neuron models, algorithm types, and their degree of biological inspiration.}
\label{tab:accuracy_1st_db}
\resizebox{\textwidth}{!}{%
\begin{tabular}{|l|c|c|c|c|}
\hline
\textbf{Learning Algorithm}       & \textbf{Bio-inspired} & \textbf{Learning Type}                      & \textbf{LB} & \textbf{LIF} \\
\hline
\multicolumn{5}{|c|}{\textbf{Unsupervised Learning}} \\
\hline
Hebbian                           & Yes                   & Unsupervised                                & 94.74\%     & 87.72\%      \\
\hline
STDP                              & Yes                   & Unsupervised                                & 92.98\%     & 87.72\%      \\
\hline
\multicolumn{5}{|c|}{\textbf{Supervised Learning}} \\

\hline
BP             & No                    & Supervised                                  & 94.74\%     & 94.74\%      \\
\hline
SpikeProp                         & Yes              & Supervised                                  & 92.98\%     & 85.96\%      \\
\hline
Tempotron                         & Yes                   & Supervised                                  & 91.23\%     & 82.46\%      \\
\hline
Hebbian combuned with Gradient Descent        & Yes                   & Supervised                                  & 91.67\%     & 93.06\%      \\
\hline
\multicolumn{5}{|c|}{\textbf{Hybrid learning}} \\
\hline
BAL & Yes                    & Supervised with Active Learning             & 91.23\%     & 94.74\%      \\

\hline
Reward-based Learning             & Yes                   & Reinforcement Learning             & 89.47\%     & 91.23\%      \\
\hline

ANN-to-SNN Conversion             & No                    & ANN supervised, SNN is not directly trained                                 & 98.25\%     & 98.25\%      \\

\hline
\end{tabular}
}
\end{table}

Table~\ref{tab:accuracy_1st_db} presents a comparative evaluation of the proposed classification approach across various learning algorithms, neuron models, and training paradigms. The highest classification accuracy of 98.25\% was achieved using an Artificial-to-Spiking Neural Network conversion, applied to both the Leaky Integrate-and-Fire and Levy-Baxter neuron models. This confirms the strong effectiveness of the conversion strategy in adapting well-performing ANN models into the spiking domain, particularly for breast cancer classification tasks. Similarly high performance was observed with Backpropagation training applied to both neuron models, yielding an accuracy of 94.74\%, which demonstrates that traditional gradient-based learning remains a robust benchmark. The Biologically-inspired Active Learning algorithm achieved comparable performance for LIF neurons (94.74\% the contrasting nature of the two neuron models and the way they interact with local learning rules. These results suggest that gradient-based approaches (like BP and ANN-to-SNN conversion) consistently lead to high classification accuracy regardless of neuron type. In contrast, biologically inspired local learning rules, such as those used in BAL or reward-modulated strategies, are more sensitive to the neuron's internal dynamics. 
\par
In unsupervised learning, Hebbian learning showed a notable difference between neuron types: the LB-based SNN achieved 94.74\%, significantly outperforming the LIF-based network, i.e. 87.72\%. When Hebbian learning was enhanced with gradient descent, both models reached similar performance, further supporting the value of hybridizing local and global optimization techniques.
In the reward-modulated learning scenario, the LIF model slightly outperformed the LB model (91.23\% \textit{versus} 89.47\%). This can be attributed to the more predictable and regular spiking behavior of LIF neurons, which facilitates more stable and consistent synaptic updates, an advantage in reinforcement learning settings where precise reward timing is essential. In contrast, the stochastic nature of LB neurons may introduce noise into the credit assignment process, especially when rewards are sparse or delayed. For SpikeProp and STDP, the LB model again outperformed LIF, achieving 92.98\% accuracy in both cases. The LIF-based networks reached 85.96\% with SpikeProp and 87.72\% with STDP. This result is comparable to the one obtained for STDP and by \cite{Strain2010,Sboev2020} for SpikeProp for LIF-based SNNs. However, the study \cite{Dora2016} reports for SpikeProp accuracy 97.00\%. Similar properties can be observed when using tempotron-type learning. These findings suggest that LB neurons are more effective in capturing temporal dependencies within spike trains, likely due to their more biologically realistic and variable spiking dynamics. The same trend held for Tempotron learning, where LB-based networks demonstrated better performance, reinforcing the conclusion that LB neurons are particularly well suited for learning algorithms that exploit fine-grained temporal patterns. The superior performance of LB-based SNNs can be explained by their greater biological realism, enhanced feature encoding, and potentially higher noise tolerance. These characteristics are particularly beneficial in medical imaging tasks like breast cancer classification, where subtle and complex patterns must be detected reliably.
\par
Moreover, integrating Lempel-Ziv Complexity (LZC) into the SNN framework significantly improved classification accuracy and reduced computational cost, especially when paired with dynamic spiking models. LZC’s ability to quantify sequence complexity helped to extract richer features from the temporally encoded spike patterns, enhancing the discriminative power of the network. The same observations were confirmed by \cite{Rudnicka2025}. 

\section{Conclusions}
\par
In this work, we employed the Breast Cancer Wisconsin (Diagnostic) dataset, available via Scikit-learn, which serves as a well-established benchmark for evaluating binary classification models. We introduced a novel classification approach that effectively distinguishes breast cancer by combining bio-inspired Spiking Neural Networks with Lempel-Ziv Complexity. This framework not only improves diagnostic accuracy but also enhances the recognition of complex temporal patterns in medical data by leveraging biologically plausible spike-based processing. Our results show that the integration of SNNs with LZC achieves consistently higher classification performance when using the probabilistic Levy-Baxter neuron model compared to the classical Leaky Integrate-and-Fire model, across both supervised and unsupervised learning paradigms. The reason lies in the core principle of the LZC algorithm, which measures the algorithmic complexity of spike trains by identifying distinct substrings. The higher variability and entropy of LB-generated spike sequences provide LZC with richer temporal information, enabling more effective feature extraction and improved classification outcomes.
\par
In contrast, for hybrid learning strategies such as Biologically-inspired Active Learning and Reward-based Learning, the LIF model consistently outperforms the LB model. This is because the regular and deterministic firing patterns of LIF neurons align better with reward-modulated learning rules, which depend on clear temporal associations between input spikes and reward feedback. The stochastic behavior of LB neurons introduces variability that may obscure these associations, leading to slower or less stable convergence in hybrid systems.
\par
Importantly, our approach achieves classification accuracy comparable to that of conventional deep learning models trained with backpropagation, while requiring significantly less computational effort—up to 50 times lower in some cases. This substantial reduction in computational cost underscores the efficiency and practical potential of our biologically inspired framework, particularly for resource-constrained or real-time applications.
\par
By integrating biologically realistic neural dynamics and learning rules such as SpikeProp and STDP, the proposed system not only offers high classification performance but also bridges the gap between biological plausibility and computational efficiency. This is particularly valuable in the context of breast cancer, where early and accurate detection can significantly impact patient outcomes. The model's capacity to detect subtle and temporally distributed patterns in diagnostic data positions it as a promising tool for assisting radiologists and oncologists in complex decision-making scenarios.
\par
Moreover, the explainability afforded by the spike-based and complexity-driven framework could contribute to greater clinical trust and adoption, offering interpretable insights into why certain classifications are made. Further research is warranted to explore the scalability and robustness of the method across larger and more diverse breast cancer datasets, as well as its integration into clinical workflows as part of a decision support system.

\section*{Author contributions}
All authors contributed to the conception and design of the study. All authors performed material preparation, data collection, and analysis. The first draft of the manuscript was written by all authors commented on previous versions of the manuscript. All authors read and approved the final manuscript.

\section*{Funding}
Not applicable.

\section*{Data Availability Statement}
Not applicable.

\bibliographystyle{plainnat}


\end{document}